\documentclass{INTERSPEECH2023}

\interspeechcameraready

\title{Mispronunciation detection using self-supervised speech representations}
\name{Jazmín Vidal$^{1,2}$, Pablo Riera$^{1,2}$, Luciana Ferrer$^2$}
\address{
  $^1$Departamento de Computaci\'{o}n, FCEyN, Universidad de Buenos Aires (UBA), Argentina\\
$^2$Instituto de Investigaci\'{o}n en Ciencias de la Computaci\'{o}n (ICC), CONICET-UBA, Argentina}
\email{\{jvidal,priera,lferrer\}@dc.uba.ar}

\begin{document}

\maketitle

\begin{abstract}
In recent years, self-supervised learning (SSL) models have produced promising results in a variety of speech-processing tasks, especially in contexts of data scarcity. In this paper, we study the use of SSL models for the task of mispronunciation detection for second language learners. We compare two downstream approaches: 1) training the model for phone recognition (PR) using native English data, and 2) training a model directly for the target task using non-native English data. We compare the performance of these two approaches for various SSL representations as well as a representation extracted from a traditional DNN-based speech recognition model. We evaluate the models on L2Arctic and EpaDB, two datasets of non-native speech annotated with pronunciation labels at the phone level. Overall, we find that using a downstream model trained for the target task gives the best performance and that most upstream models perform similarly for the task. 

\end{abstract}
\noindent\textbf{Index Terms}: computer-assisted language learning, mispronunciation detection, self-supervised learning, low-resources

\section{Introduction}
\label{sec:intro}

Computer-aided pronunciation training (CAPT) systems provide feedback to second language learners on their pronunciation quality, with positive impacts on learning and motivation~\cite{rogerson2021computer}.
One family of CAPT systems frames the problem as a phone recognition task, using non-native data during training \cite{li2016mispronunciation, leung2019cnn, peng2021study}. These systems identify pronunciation errors by comparing the phonetic transcription of a student's speech to a native target sequence using dynamic programming algorithms. %Then, they provide feedback using information about the mispronounced phone variant. 
Another family of CAPT systems frames the problem as detection of mispronunciations, generating scores that are then thresholded for the final decision. These systems can be classified into two groups. Those that do not use non-native data during training rely on automatic speech recognition (ASR) systems trained with native speakers, and generate pronunciation scores using the acoustic model's outputs \cite{franco1997automatic, witt2000phone, hu2015improved, shi2020context}. The most widely used approach in this family is called Goodness of Pronunciation (GOP) \cite{witt2000phone}. The second group uses non-native data to directly train the system to distinguish correctly- from incorrectly-pronounced segments using a variety of input features and classifiers \cite{ito05_interspeech, truong2004automatic, van2009automatic, yoon2010landmark, franco2014adaptive}. Recently, transfer learning techniques have been used to mitigate the problem of data scarcity that is the norm in the task. In these approaches, deep neural networks (DNNs) models trained for ASR  \cite{hu2015improved, huang2017transfer, sancinetti2022transfer, gong2022transformer} or on a self-supervised fashion \cite{xu2021explore} are fine-tuned to detect mispronunciations. 

In this work, we target systems of the second family described above. These systems have the advantage of providing a measure of the confidence that they have in their detection, enabling adjustment of false correction rates to acceptable levels to avoid frustrating the student in real educational scenarios~\cite{neri2002pedagogy}. We explore two different approaches for using representations obtained from pre-trained self-supervised or supervised models. In both cases, the representations are fed to a downstream model which generates scores for each target phone. In the first approach, the downstream model is trained for the task of phone recognition (PR) using native data only. In the second approach, the system is trained for the mispronunciation detection (MD) task using non-native speech datasets annotated with pronunciation labels at the phone level. The latter approach is similar to the one in \cite{xu2021explore}, where authors explore the use of Wav2vec 2.0 \cite{baevski2020wav2vec} for the L2-Arctic database, a database of non-native English speakers of many L1 backgrounds. Unfortunately, their code is not publicly available. 

Our work aims to explore the approach in \cite{xu2021explore} in further detail across a variety of SSL models and to compare it with the PR approach where downstream model training does not require non-native annotated data. To this end, we show results using various self-supervised models (WavLM+, WavLM Large, HuBERT and LightHuBERT Small) and one supervised model (TDNN-F) to extract representations. 
For evaluation we use two publicly available databases: L2-Arctic \cite{zhao2018l2} and EpaDB \cite{vidal2019epadb}, a database of non-native English speakers from Argentina.
For performance assessment, we use the Area Under the ROC curve (AUC) and a metric designed to encourage low false correction rates, proposed in our prior work \cite{vidal2021phone, sancinetti2022transfer}. We release the code to reproduce the experiments as a s3prl \cite{s3prl} recipe (temporary URL \url{https://github.com/JazminVidal/ssl-mispron}). 

\section{Upstream models}
\label{sec:ssl}
In this work, we study the use of pre-trained SSL models for the task of mispronunciation detection. Self-supervised learning has emerged as a promising approach for speech representation learning since it can leverage large amounts of unlabelled speech data to produce effective representations. %We use the models as off-the-shelf feature extractors to reflect the practical reality of the task: a low-resource scenario in which training large models is infeasible.

One of the first SSL speech models to have a large impact on various downstream tasks was Wav2vec2 2.0 \cite{baevski2020wav2vec}. Following its steps, many new models appeared in recent literature. Among these, we select HuBERT \cite{hsu2021hubert} and WavLM+ \cite{wavlm:2022} because they have a good performance on the phone recognition task in the Speech Processing Universal Performance (SUPERB) benchmark \cite{yang2021superb}. We also try a larger and a smaller model: WavLM Large \cite{wavlm:2022} and LightHuBERT Small \cite{wang2022lighthubert} to explore the effect of model size.  

Two essential components comprise all of these models: a CNN encoder and 12 to 24 transformer layers. The models are trained to predict masked targets using the transformer's output. This encourages the model to leverage long-term dependencies in the speech signal, which are essential for the task of ASR and others. In HuBERT, during the first training stage, the targets are generated by k-means clustering of Mel Frequency Cepstral Coefficients (MFCCs). In the final stage, the MFCCs are replaced with latent variables from the same model, allowing it to perform iterative refinement of the target labels. The model uses a similarity loss function between the target and the predictions. WavLM uses a similar strategy to HuBERT, but the training data is generated as mixtures of speech and noise. The objective is to predict the target clean speech signal from a noisy masked one, which makes the system more robust. 
In this work, we also use the WavLM+ (``Plus'') version that was trained with +90k hours of speech, compared with the 960 hours used for training the HuBERT and the WavLM models. LightHuBERT is a distilled version of HuBERT which has similar performance but half of the parameters. All of these models have 12 transformer layers, except for WavLM Large which has 24 layers. The size of the activations is 384 for LightHuBERT, 768 for HuBERT and WavLM+ and 1024 for WavLM Large. All models have a frame rate of 50HZ. 

The traditional ASR system we use for comparison is the Kaldi Librispeech ASR model, a TDNN-F \cite{povey2018semi} acoustic model trained for the task of senone recognition. This model consists of 18 hidden layers (some factorized, some with time delay) with ReLU activations and skip connections. The last hidden layer is linear and has an output dimension of 256. The output layer of this model consists of 6024 nodes, one per senone. The model was trained with 960 hours of native English speech from the LibriSpeech \cite{panayotov2015librispeech} dataset. 

\begin{figure*}[ht!]
\begin{centering}
\includegraphics[width=0.75\textwidth]{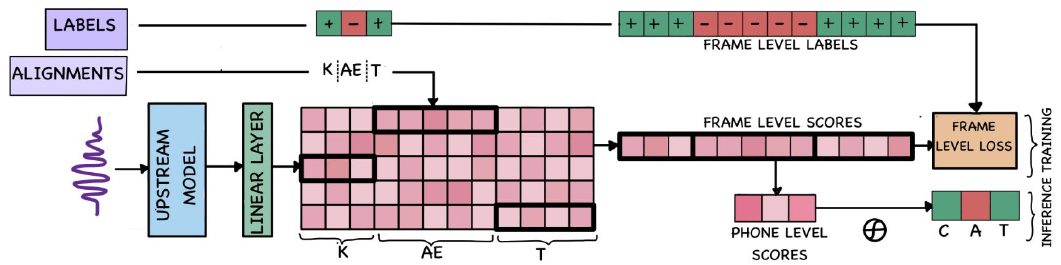}
\vspace{-2mm}
\caption{Schematic of our approach. Frame-level outputs of an upstream model are fed to a linear layer that produces one score per target phone in the English language and per frame in the phrase. Next, using the time alignments provided as input, we select the scores corresponding to the target phones detected at each frame. Finally, phone-level scores are computed by averaging over all the frames for each of these targets. For the final classification into correct or incorrect pronunciations, we compare the phone-level scores to a threshold tuned on the development data for each phone. For MD downstream models the linear layer is trained using the frame-level loss shown in this figure which uses the correct versus incorrect labels for each target phone in the alignment, replicating them over all frames in each phone. PR downstream models are instead trained for the task of phone recognition.}
  \label{fig:fig_method}
  \end{centering}
\end{figure*}

\section{Downstream models}
\label{sec:downstream}
In this section, we describe the two approaches that we explore for training the downstream models, the phone recognition (PR) approach and the mispronunciation detection (MD) approach. Figure \ref{fig:fig_method} shows a schematic of the proposed model. In both cases, the downstream model is a linear layer with one node per English phone. The layer takes the representations from the upstream models as input. What differs between the approaches is the way the parameters of this linear layer are learned. Further, both approaches require phone-level time alignments as input. The alignments are obtained with an automatic forced-aligner (described in Section \ref{sec:preproc}) and indicate the location of each phone in the transcription. 

In the {\bf PR approach}, for the SSL models, the downstream model is trained for phone classification using a dataset of native English speakers. The output layer of this downstream model has a softmax activation, generating the posterior probabilities of each English phone for each frame. For the TDNN-F model, no downstream model is trained. Instead, the output layer of the model is used directly to compute the per-phone scores by averaging the posteriors for all senones corresponding to each phone. This combination of TDNN-F upstream model followed by the PR approach coincides exactly with the standard GOP algorithm, as implemented in the Kaldi recipe and shared in \cite{sancinetti2022transfer}, which we take as our baseline. 

In the {\bf MD approach}, we use annotated non-native English data to train the downstream model to directly detect correctly versus incorrectly pronounced phones. In this case, the output layer of the model has a sigmoid activation in each node rather than a softmax activation. Again, each node corresponds to an English phone, but its value is now meant to predict whether the corresponding target phone was correctly or incorrectly pronounced. The training loss is computed by selecting the output node corresponding to the target phone pronounced at each frame (according to externally-provided alignments) and then averaging the loss over all selected scores.

The input to the downstream models for the SSL upstream models is given by a weighted sum of the encoder activations plus the activations of the transformer layers (24 layers for WavLM Large, 12 for the rest of the models). The weights are learned along with the linear layer parameters of the downstream model. We train the downstream in the TDNN-F model for the MD approach following the methods and code proposed in \cite{sancinetti2022transfer} where the output layer is stripped from the TDNN-F model and the 256-dimensional layer is fed as input to the linear layer. 
Finally, once the downstreams are trained, scores are computed using their frame-level outputs before the activation function (softmax for PR, sigmoid for MD) is applied. For a target phone $p$ that starts at frame $T$ and has a length of $D$ frames the score is computed as $ \mathrm{Score}(p)=\frac{1}{D}\sum_{t=T}^{T+D-1} s_{t,p}$
where $s_{t,p}$ is the pre-activation output at frame $t$ for phone $p$.

\section{Experimental setup}
In this section, we describe the databases used in our experiments, the audio pre-processing, the model training procedure and the evaluation metrics.
\subsection{Databases}
\label{sec:dbs}
The TIMIT dataset was used for training the PR downstream models, while Epa-DB and L2Arctic were used for MD model training and final evaluation of the approaches. 

\textbf{TIMIT} \cite{timit:1993} is a database of read native English speech designed for the development and evaluation of ASR systems. It contains a total of 6300 sentences recorded at 16kHz by 630 speakers of 8 dialects of American English. Each speaker recorded 10 phonetically rich sentences. The database includes  time-aligned orthographic, phonetic and word transcriptions. 

\textbf{Epa-DB} \cite{vidal2019epadb} is a database of non-native English speech by Spanish speakers from Argentina. Recording may contain noise and vary in sample rate between 16kHZ and 44kHZ. The database contains 3200 English short utterances produced by 50 speakers (25 male and 25 female). Manual annotations of pronunciation labels are included for each audio sample.

\textbf{L2-ARCTIC} \cite{zhao2018l2} is a database of non-native English speech intended for research in accent conversion and mispronunciation detection. The speech was recorded in a controlled scenario, under quiet conditions using quality microphones with a sample rate of 44.1kHz. There are 3621 annotated recordings from 24 nonnative speakers (12 males and 12 females) whose first languages (L1s) are Hindi, Korean, Mandarin, Spanish, Arabic and Vietnamese. A subset of 150 utterances per speaker is manually annotated with mispronunciation errors. 

\subsection{Preprocessing and Model Training}
\label{sec:preproc}
For this work, the set of target phones is given by the 39 phones in the English language plus silence \cite{TIMITphones}. To annotate the mispronunciations, L2Arctic uses International Phonetic Alphabet (IPA) symbols, whereas EpaDB and TIMIT use ARPAbet coding. To normalize annotations, we mapped L2Arctic phones to ARPAbet. Further, TIMIT included two extra phones not considered in the other two databases, /AX/ and /DX/. We mapped them to /AH/ and /T/, their closest phones in our phone set. 

As explained in Section \ref{sec:downstream}, our models require alignments and labels at the frame level. We obtain time-alignments and phonetic transcriptions using a Kaldi TDNN-F based forced aligner implemented in PyKaldi \cite{pykaldi}, the same model we use on the PR and MD supervised systems. We take the phonetic transcriptions returned by the forced aligner as the targets the student should have pronounced. These are our individual samples for prediction. The total count of target phone instances per database is 39775 positives and 10480 negatives for EpaDB and 99799 positives and 18539 negatives for L2Arctic. TIMIT contains 241225 native phone instances.

For training of MD models and evaluation, we align each sequence of target phones with its corresponding sequence of manual annotations and assign their labels. Positive labels are assigned to correctly pronounced phones and negative labels to incorrectly pronounced ones. For L2-Arctic we use a version of the ALINE algorithm \cite{kondrak2000new} adapted to work on ARPAbet symbols. For EpaDB we follow the alignment scripts provided in \cite{sancinetti2022transfer}. 
Note that if a non-native sample contains an error where one or more phones are added between two target phones, one must decide which target to assign the negative label that corresponds to the added phone(s). For simplification, we follow the approach taken in \cite{zhao2018l2} and ignore all addition errors marked in L2-Arctic. This is not necessary for EpaDB because additions in this database are annotated as a substitution of the target phone by two new phones. 

Next, we down sample all the waveforms in EpaDB and L2Arctic to 16kHz and partition the three databases into subsets. For TIMIT we use the training, development and testing splits specified in the s3prl phone recognition recipe. For EpaDB, we assign 30 speakers for development and 20 for testing. For L2Arctic, we use the 20 speakers with non-Spanish L1 backgrounds for development and leave the 4 Spanish-L1 background speakers for testing, to make it comparable to the EpaDB test set.
We train the PR models on the training split of TIMIT and select the best model using the development split. Also, in the PR case, the development split of the non-native dataset being evaluated is used for selecting the decision threshold on the test data. For the MD models, we first obtain scores in the full development set by doing K-fold cross-validation. The pooled scores are used to determine the decision threshold. For EpaDB we use 6 folds, divided by speaker. For L2Arctic we use 5 folds, divided by L1. Finally, for evaluation of the test split, we train a model using the full development set. Splits and fold lists for each database and system are provided with the code in the repository listed in Section \ref{sec:intro}. 
 
We implement the SSL models using \emph{s3prl}. For the downstream models for the SSL case, we use AdamW optimization and mini-batches of 64 samples using a learning rate of $1\times10^{-4}$ ($1\times10^{-5}$ for Large model). The best number of training epochs was selected based on cross-validation results. For the TDNN-F model, we follow the implementation provided in \cite{sancinetti2022transfer} where we train the last layer using the Adam optimizer, with mini-batches of 32 samples over 300 epochs using a learning rate decaying every 10 epochs by a factor of 0.9, starting from 0.01. Experiments were done using an RTX 3090 GPU. 

\subsection{Evaluation metrics}
\label{sec:metrics}
Our systems generate scores for each target phone which are expected to have higher values for correctly pronounced phones than for incorrectly pronounced ones. Hard decisions can then be made by comparing these scores with a threshold. Each possible threshold results in a false positive rate (FPR) and false negative rate (FNR). 
In our results, we report the area under the false negative versus false positive rate curve (equivalent to 1 minus the traditional AUC metric). This metric integrates the performance over all possible operating points given by different thresholds and is a very standard metric used for this and many other tasks.
In addition, following \cite{sancinetti2022transfer}, we report another metric considered to be more appropriate for the task of mispronunciation detection, where the false negative rate should be minimized to avoid frustrating the student with unnecessary corrections. We define a cost given by  FPR + 2 FNR, where FNR is penalized more than the FPR, prioritizing low FNR over low FPR. This type of cost function is widely used in speaker verification and language detection tasks \cite{van2007introduction}, where the weights are determined depending on the application scenario.
To compute the cost, a decision threshold is needed. One possible approach is to choose the threshold that minimizes the cost for each phone on the test data itself, resulting on the best possible cost on that data (MinCost). Selecting the optimal threshold on the test data, though, leads to optimistic estimates of the cost. Hence, for the test data, we also compute the cost obtained when the threshold is selected as the one that optimizes the cost of the development data for each phone. We call this the Actual Cost (ActCost). 

Both 1-AUC and costs are computed as averages over phones. This is because if all samples are pooled together for computation of the metrics, the most frequent phones  dominate the value of the metric. Yet, for this application, infrequent phones are equally important as frequent ones. Further, since some phones in EpaDB and L2-Arctic have very few instances of incorrect pronunciation and the performance on these phones cannot be robustly estimated, when computing the average metrics we discard all phones with less than 50 instances of the minority class. Finally, we use bootstrapping \cite{poh2007bootstrap} with speaker-level sampling to compute confidence intervals and assess statistical significance of differences between systems.

\begin{figure}[t]
\centering
\includegraphics[scale=0.4]{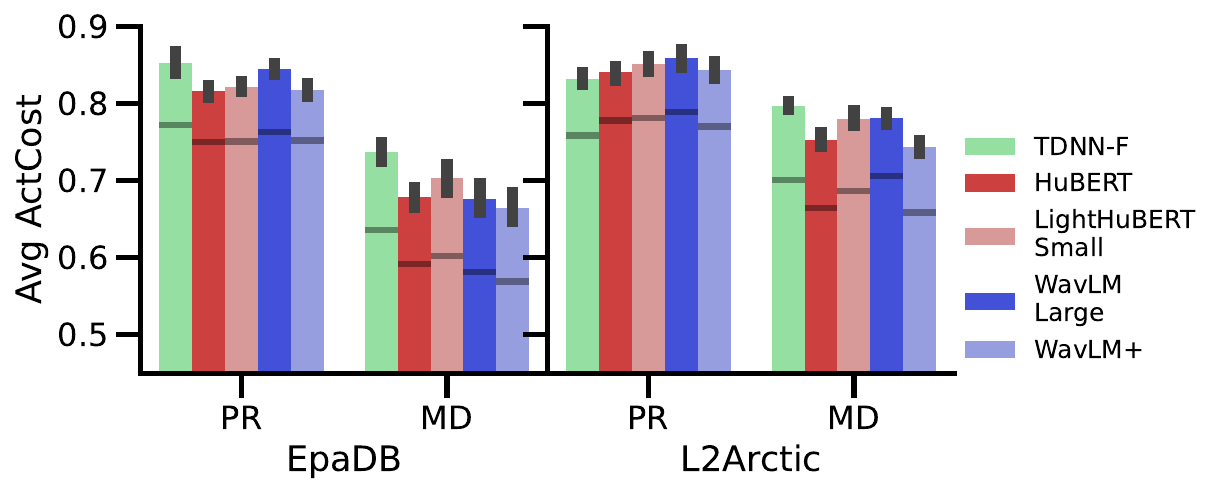}
\vspace{-7mm}
\caption{Results for EpaDB and L2-Arctic comparing different upstream (legend) and downstream (x-label) combinations. Values correspond to the average ActCost over phones with more than 50 samples of each class for the test data. The solid lines in each bar show the average MinCost, the optimal cost for the test data. Vertical lines indicate bootstrap confidence intervals.}
  \label{fig:fig_res}
\end{figure}

\section{Results}
\label{sec:res}

Figure \ref{fig:fig_res} shows the average ActCost and MinCost for the test splits of EpaDB and L2-ARCTIC. Each group of bars compares different systems, for the phone recognition (PR) and mispronunciation (MD) downstream approaches. Comparing the PR and the MD groups, we see that training with non-native data to directly detect mispronunciations results in better performance than using the outputs of a phone recognizer to compute mispronunciation scores. Yet, in scenarios where little labeled non-native data is available, the MD approach may not be feasible, while the PR approach can still be used. 

Importantly, the figure shows that even for the PR systems the average ActCost is significantly better than 1, meaning the systems are all better than the best naive system. This indicates that such systems would still be useful in practice according to this application-motivated metric. Notably, all SSL and the TDNN-F upstream models lead to similar performance values for each of the two downstream approaches. The relatively small differences between models' performances show that the Large WavLM model is not the best for this task, despite being the best model for the native phone recognition task in the SUPERB benchmark. Also, the Small LightHuBERT model is not far behind the rest of the models, showing similar performance to HuBERT which has twice as many parameters. Again, while HuBERT is better than LightHuBERT for the task of phone recognition according to the SUPERB benchmark, this advantage does not carry over to the task of mispronunciation detection. Overall, the WavLM+ model gives the best or close to best results across both datasets.

The solid horizontal lines inside the bars indicate the MinCost. This is an optimistic estimate of the cost, since, in practice, one never has the full evaluation data on which to estimate the thresholds. The height of the bars on the other hand, is the ActCost, computed using the threshold estimated on the development data. We can see that the ActCost is within 10\% of the MinCost for most PR systems, both for EpaDB and Arctic, indicating that the threshold can be robustly selected based on the development data. For the MD results, the discrepancy is a little larger, but it does not exceed 15\%.

{
\begin{table}[t]
\footnotesize
\centering
\begin{tabular}{p{3pt}lrrrr}
 & & \multicolumn{2}{c}{EpaDB} & \multicolumn{2}{c}{L2-Arctic} \\
 & & 1-AUC & ActCost & 1-AUC & ActCost \\
\midrule
PR & TDNN-F      & 0.29 & 0.85  & 0.29 & 0.83 \\
 & WavLM+        & 0.33 & 0.82  & 0.33 & 0.84 \\
\midrule
MD & TDNN-F     & 0.2 & 0.73  & 0.24 & 0.79 \\
& WavLM+        & 0.17 & 0.67  & 0.17 & 0.74 \\
\bottomrule
\end{tabular}
\caption{Average ActCost and 1-AUC, for PR and MD approaches and both datasets for TDNN-F and the best SSL model WavLM+. The first line, PR-TDNN-F, corresponds to the GOP baseline.}
\label{tab:my_label}
\end{table}
}
Table \ref{tab:my_label} shows 1-AUC and ActCost for the PR and the MD approaches for two models: TDNN-F which can be considered a baseline, and WavLM+. We can see a discrepancy between 1-AUC and ActCost both in terms of trends (a system may be better than another for one metric but worse for the other metric) and in terms of absolute values (1-AUC are relatively farther from their baseline, which is 0.5, than ActCost values from their baseline, which is 1.0). Two things explain these discrepancies. First, 1-AUC ignores the problem of threshold selection which is equivalent to assuming that the threshold will always be set optimally - an unreasonable assumption. On the other hand, the ActCost values are affected by the threshold selection process. Second, 1-AUC integrates the performance of the whole range of operating points, while the cost focuses on a single point, which is the point of interest for this application. Given this discrepancy, and the fact that 1-AUC is not directly reflecting the performance of interest for our task (both because it ignores the threshold selection problem and because it does not focus on a relevant operating point), we believe 1-AUC and other metrics that suffer from one or both these issues, like the F1-score, are not appropriate for evaluation of this task. For a related discussion on classification metrics, see \cite{ferrer2022analysis}. 

Additionally, we run a series of experiments to explore potential improvements in performance though, unfortunately, did not find any significant gains. We explored different training losses with phone-level summarization and class weighting in the loss. We tried more complex downstream architectures, such as CNNs, but found that they tended to overfit more than simpler linear models. We also tried fine-tuning some upstream layers but the results were comparable to those obtained without fine-tuning, although further exploration of training hyper-parameters should be performed. 

\section{Conclusions}
In this study, we explored addressing the mispronunciation detection task by using pre-trained self-supervised (SS) and supervised models to generate speech representations which are then used as input to downstream models for score generation. We compared two approaches for training the downstream models, using only native data and using non-native data annotated for pronunciation scoring, finding that, as expected, the latter approach leads to improved performance over the first one. The first approach, though, can be used when not enough annotated non-native data is available for model training. Among the SS models tested, WavLM+ achieved the best performance, followed closely by all other models, indicating that the specific upstream model used has a much smaller effect than the downstream approach. 

%\section{Acknowledgements}
%This project received funding from the European Union’s Horizon 2020 research and innovation program under grant No 101007666/ESPERANTO/H2020-MSCA-RISE-2020.

\bibliographystyle{IEEEtran}
\bibliography{mybib}

\end{document}